\documentclass{article}

\PassOptionsToPackage{numbers, sort}{natbib}
 \usepackage[preprint]{neurips_2026}

\usepackage[utf8]{inputenc}
\usepackage[T1]{fontenc}
\usepackage{url}            
\usepackage{amsfonts}
\usepackage{nicefrac}
\usepackage{microtype}
\usepackage{xcolor}
\usepackage{graphicx}
\usepackage{colortbl}
\usepackage{makecell}
\usepackage{tabularray}
\usepackage{multirow}
\usepackage{kotex}
\usepackage{amsmath}
\usepackage{wrapfig}
\usepackage{booktabs}
\usepackage{bbm}

\usepackage[citecolor=blue, colorlinks]{hyperref}

\definecolor{lightgray}{gray}{0.92}

\newcommand{\myparagraph}[1]{\vspace{1pt}\noindent{\bf #1}}

\definecolor{caribbeangreen}{rgb}{0.0, 0.8, 0.6}
\definecolor{amethyst}{rgb}{0.6, 0.4, 0.8}
    
\title{Causality-Aware End-to-End Autonomous Driving \\via Ego-Centric Joint Scene Modeling}

\author{Seokha Moon$^{1}$, Minseung Lee$^{1}$, Joon Seo$^{1}$, Jinkyu Kim$^{1,2}$, Jungbeom Lee$^{1}$\\
$^1$Korea University, $^2$Kakao Mobility}

\begin{document}

\maketitle

\begin{abstract}
End-to-end autonomous driving, which bypasses traditional modular pipelines by directly predicting future trajectories from sensor inputs, has recently achieved substantial progress. However, existing methods often overlook the causal inter-dependencies in ego-vehicle planning, ignoring the reciprocal relations between the ego vehicle and surrounding agents. This causal oversight leads to inconsistent and unreliable trajectory predictions, especially in interaction-critical scenarios where ego decisions and neighboring agent behaviors must be reasoned about jointly. To address this limitation, we propose \textbf{CaAD}, a Causality-aware end-to-end Autonomous Driving framework that captures these dependencies within a shared latent scene representation. First, we propose a ego-centric joint-causal modeling module that builds on the marginal prediction branch, and learns causal dependencies between the ego vehicle and interaction-relevant agents. Second, we employ a causality-aware policy alignment stage implemented with joint-mode embeddings to align the stochastic ego policy with planning-oriented closed-loop feedback computed from surrounding traffic and map context. On the Bench2Drive and NAVSIM benchmarks, CaAD demonstrates strong closed-loop planning performance, achieving a \textbf{Driving Score of 87.53} and \textbf{Success Rate of 71.81} on Bench2Drive, and a \textbf{PDMS of 91.1} on NAVSIM. The project page is available at \url{https://moonseokha.github.io/CaAD/}.
\end{abstract}

\section{Introduction}
\label{sec:intro}
End-to-end (E2E) autonomous driving has recently emerged as a promising alternative to traditional modular pipelines~\cite{voxelnet, bevfusion, bevformer, qcnet, visiontrap, centerpoint, second, scenetransformer, e2e}, directly predicting future trajectories or control actions from sensor inputs~\cite{chen2024end,transfuser,chitta2021neat,uniad,vad,sparsedrive}. 
By learning a unified mapping from perception to planning, E2E models reduce the reliance on hand-crafted intermediate representations and mitigate error accumulation across separately optimized stages. With the availability of large-scale driving datasets and closed-loop benchmarks~\cite{nuscenes,nuplan,b2d,navsim}, most recent E2E planners are trained primarily with imitation learning (IL), where the model is optimized to reproduce expert driving behavior~\cite{transfuser,chitta2021neat,uniad,vad,vadv2,hipad}. This paradigm has led to substantial progress, demonstrating the effectiveness of data-driven planners in learning complex driving policies from expert behaviors.

\begin{figure}[t!]
  \centering
  \includegraphics[width=1\linewidth]{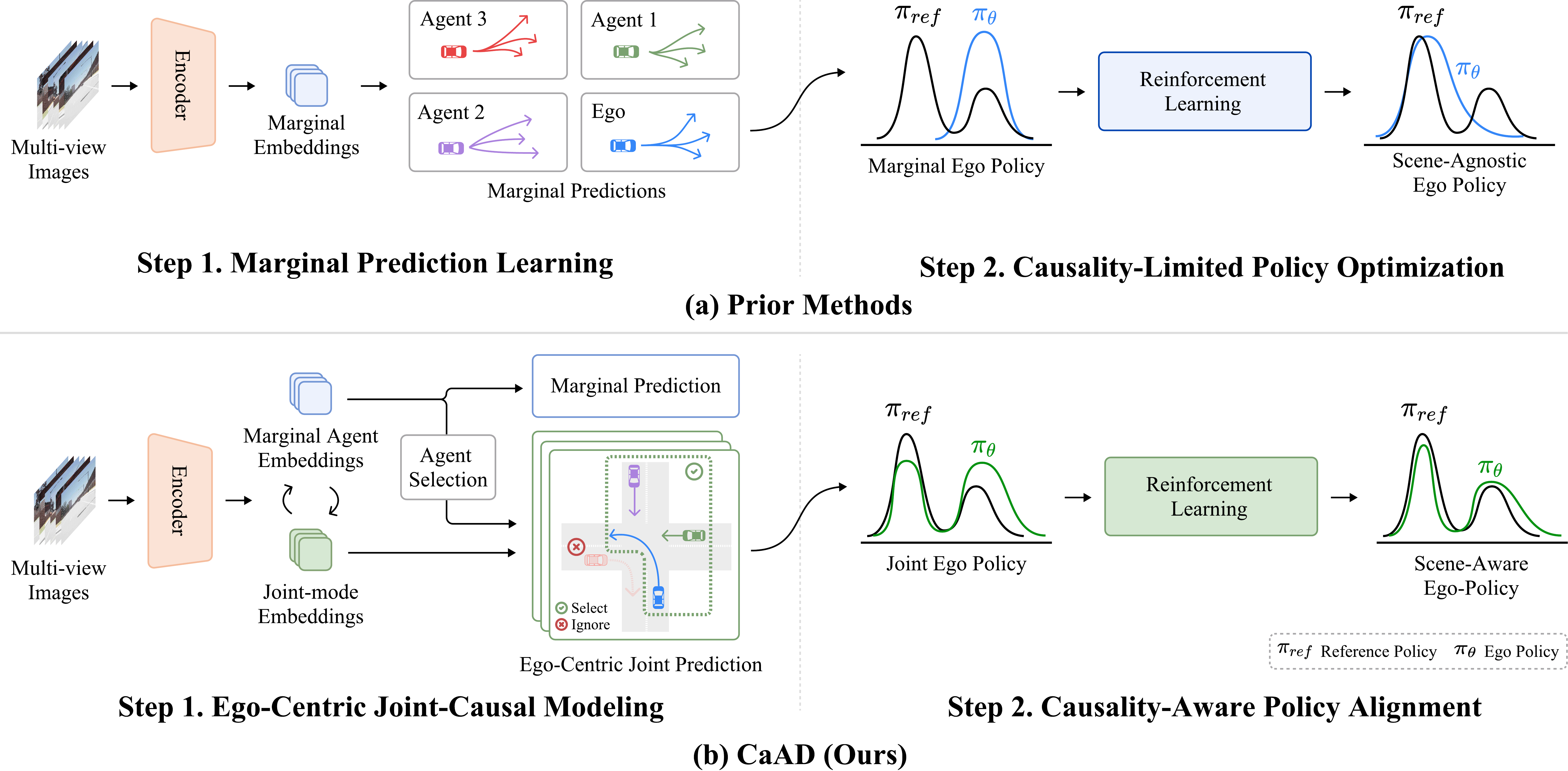}
  \vspace{-1em}
  \caption{Comparison between prior E2E methods and CaAD. (a) Prior methods first perform marginal prediction learning and then optionally adopt RL-based policy optimization on marginal representations, resulting in causality-limited policy refinement. (b) We first performs ego-centric joint-causal modeling and then applies causality-aware policy alignment on joint-mode embeddings, where each sampled ego rollout is evaluated.}
\label{fig:teaser}
\vspace{-1.5em}
\end{figure}

Despite this progress, existing E2E methods still struggle to represent mutually consistent futures of the ego vehicle and surrounding agents in highly interactive traffic, largely because they model all agents marginally (Step 1 of Fig.~\ref{fig:teaser}(a)). Reliable driving requires more than predicting an individually plausible ego trajectory; the predicted ego motion must remain coherent with the future behavior of nearby agents. In real-world driving, a merge may succeed only if a neighboring vehicle yields, while an overtake may be safe only when surrounding vehicles maintain compatible motions. Similarly, crossing an intersection may depend on the behavior of conflicting traffic. Such interaction dependencies have been extensively studied in the modular motion forecasting task, where independent marginal futures are known to be insufficient for faithfully representing multi-agent scenes~\cite{scenetransformer,sun2022m2i,seff2023motionlm,wagner2024jointmotion,wayformer}. In contrast, prior E2E driving methods still predict the ego plan and surrounding-agent motion as loosely coupled outputs, relying mainly on shared latent features to capture interactions implicitly~\cite{uniad,vad,vadv2,sparsedrive,hipad}. Consequently, trajectories that are reasonable in isolation may form a globally inconsistent traffic scene, resulting in unreliable closed-loop behavior during critical interactions like merging or yielding.

To address this limitation, we propose \textbf{CaAD}, a Causality-aware end-to-end Autonomous Driving framework that captures these dependencies by organizing shared features into ego-conditioned joint scene representations rather than using shared latent context alone. As shown by Step 1 in Fig.~\ref{fig:teaser}(b), CaAD performs ego-centric joint-causal modeling through interaction between marginal agent embeddings and joint-mode embeddings, where marginal embeddings retain actor-specific motion evidence and joint-mode embeddings encode ego-centric scene hypotheses for modeling interactions. This ego-centric formulation is well suited for planning-oriented scene modeling, where uncertainty is primarily driven by the subset of agents critical to the ego-vehicle's safety and feasibility. The resulting interaction yields joint-mode representations that facilitate coupled prediction of the ego-plan and relevant agent behaviors. This enables the planner to reason over interactive futures while maintaining the computational efficiency of planning-oriented E2E architectures.

Supervised imitation alone does not directly optimize closed-loop driving quality, since pointwise trajectory losses may underweight key objectives such as safety, map compliance, comfort, and long-horizon progress~\cite{recogdrive,emma}. Accordingly, although some methods~\cite{diver,recogdrive,shang2025drivedpo} further refine the ego policy with reward or preference signals, such refinement remains causality-limited, as it is still built on marginal or ego-focused planning representations. In contrast, as shown in Step 2 of Fig.~\ref{fig:teaser}(b), our method performs causality-aware policy alignment on ego-centric joint-mode embeddings, allowing reinforcement learning to refine an ego policy that has already been shaped by joint-causal modeling. This keeps surrounding-agent prediction supervised while shifting the ego policy toward safer and more coherent interaction outcomes, leading to state-of-the-art performance on Bench2Drive~\cite{b2d} and NAVSIM~\cite{navsim}.

Our contributions can be summarized as follows:
\begin{itemize}
    \item \textbf{Joint-causal modeling.} To the best of our knowledge, CaAD is the first end-to-end driving framework to explicitly incorporate ego-centric joint prediction. By facilitating interactions between marginal agent embeddings and ego-centric joint-mode embeddings, CaAD constructs shared representations which enable the model to learn causality-aware scene dynamics.
    \item \textbf{Causality-aware policy alignment.} We employ a causality-aware policy alignment stage built on ego-centric, joint-mode embeddings to align a stochastic ego policy with planning-oriented, closed-loop feedback. This feedback is computed against the surrounding traffic and map context, while keeping the surrounding-agent forecasting supervised.
    \item \textbf{State-of-the-art closed-loop performance.} CaAD achieves state-of-the-art performance on both Bench2Drive and NAVSIM. On Bench2Drive, it achieves a \textbf{Driving Score of 87.53} and a \textbf{Success Rate of 71.81}. On NAVSIM, it attains the best \textbf{PDMS of 91.1}.
\end{itemize}

\section{Related Work}

\subsection{End-to-End Planning and Reinforcement Learning for Autonomous Driving}
End-to-end autonomous driving learns planning-oriented representations directly from sensor observations, offering an alternative to modular perception--prediction--planning pipelines~\cite{uniad, chen2024end,paradrive,vad,transfuser,chitta2021neat, sparsedrive, hipad}. To better capture multi-modality and uncertainty, generative planning methods have further emerged, ranging from anchor-based methods~\cite{vadv2, sparsedrive, sparsedrivev2, hipad} to diffusion- or flow-based trajectory generation~\cite{genad, diffusiondrive, goalflow}. Recent studies have also explored world-model-based planning, motivated by its ability to model future scene evolution~\cite{think2drive, raw2drive, world4drive, law, wote, seerdrive, resworld}. In parallel, reinforcement learning and preference-based optimization have been used to refine imitation-trained driving policies toward objectives beyond behavior cloning~\cite{chekroun2023gri,jiang2025alphadrive,lu2023imitation,li2025learning,shang2025drivedpo,alpamayo,recogdrive,autodrive}. However, existing approaches still primarily optimize the ego planner rather than explicitly modeling reciprocal ego--agent future dynamics. In contrast, our proposed CaAD introduces ego-centric joint prediction and a scene-level reinforcement learning stage, enabling coupled prediction of the ego and ego-relevant surrounding agents in a shared latent space.

\subsection{Joint and Interaction-Aware Motion Forecasting}
In interactive driving, independently predicted marginal futures can be mutually inconsistent, motivating joint models for scene-consistent agent futures~\cite{scenetransformer,sun2022m2i,seff2023motionlm,wagner2024jointmotion,wayformer, girgis2022autobots, shi2022mtr,luo2023jfp}. Recent methods improve interaction-aware forecasting with scene-centric~\cite{scenetransformer}, factorized~\cite{sun2022m2i}, latent-variable~\cite{girgis2022autobots}, graph-based joint future prediction~\cite{luo2023jfp}, transformer query-based prediction and refinement~\cite{shi2022mtr}, token-based language modeling~\cite{seff2023motionlm}, self-supervised pre-training~\cite{wagner2024jointmotion}, and retrocausal/instructable formulations~\cite{wagner2025retromotion}. Complementary safety-aware planners further estimate future collision risks and drivable-region compliance to support interpretable end-to-end driving~\cite{kim2026safedrive}. However, these works mainly address ego-side prediction or safety-aware ego planning, rather than explicitly optimizing a globally coordinated multi-agent plan. CaAD augments marginal forecasting with a selective joint module that captures causal dependencies between the ego vehicle and interaction-relevant actors. A more detailed discussion of earlier end-to-end planners, RL-based policy refinement, and joint forecasting models is deferred to Appendix~\ref{app:related_work_details}.
\section{Method}
\label{sec:method}

\begin{figure}[t!]
  \centering
  \includegraphics[width=1\linewidth]{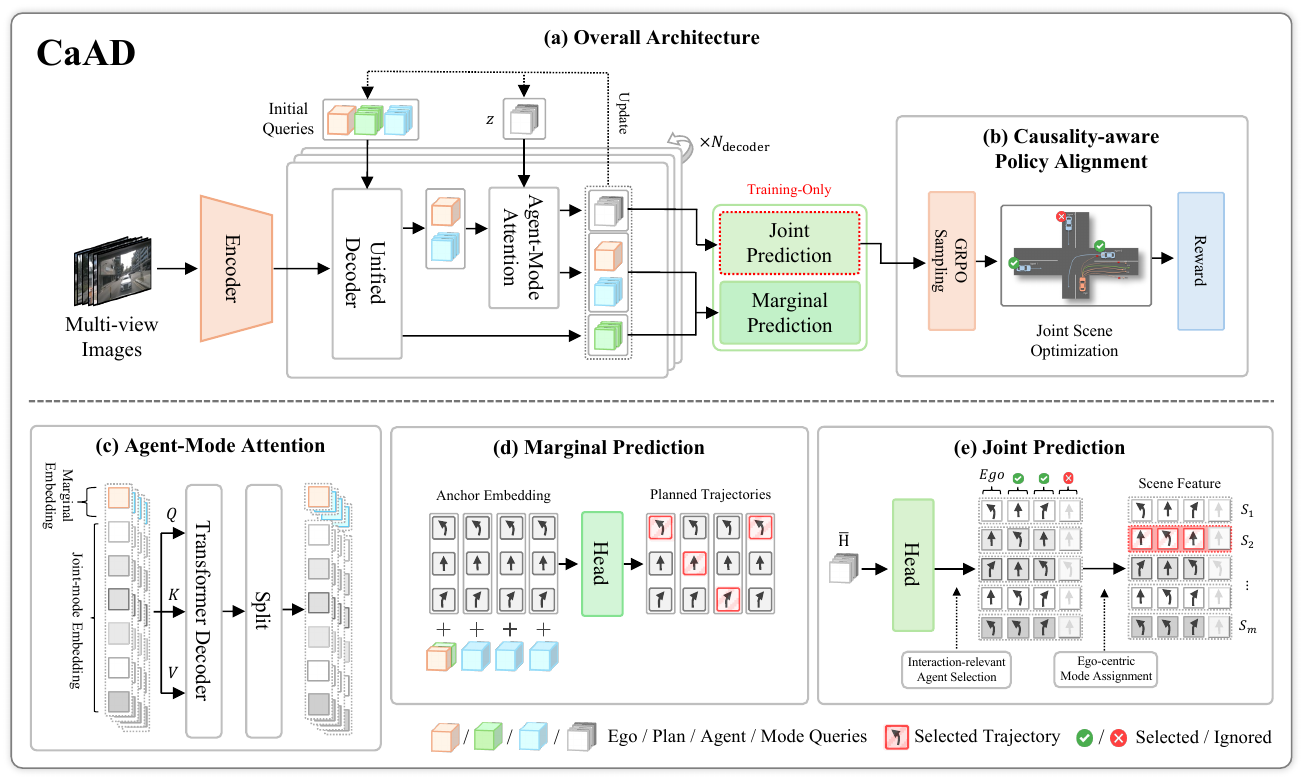}
  \vspace{-1.5em}
  \caption{Overview of the proposed CaAD. (a) CaAD builds on a query-based architecture and augments the marginal prediction branch with a training-only ego-centric joint prediction branch. (b) A causality-aware policy alignment stage further aligns the stochastic ego policy with planning-oriented rewards computed from surrounding traffic and map context. (c) Agent-Mode Attention enables marginal agent embeddings and joint-mode embeddings to interact to learn scene representations. (d) The marginal branch preserves stable trajectory prediction, (e) while the joint branch supervises interaction-relevant agents under ego-centric scene modes. 
}
\label{fig:main}
\vspace{-1.em}
\end{figure}

CaAD addresses the loose coupling between ego planning and surrounding-agent forecasting in interactive end-to-end driving, where individually plausible futures can be interaction-inconsistent. As shown in Fig.~\ref{fig:main}, CaAD augments a query-based E2E planner with ego-centric joint-causal modeling through marginal--joint embedding interaction. Marginal embeddings preserve actor-specific evidence, while ego-centric joint-mode embeddings organize predictions into shared scene modes (Sec.~\ref{sec:joint_prediction}), yielding scene representations that align the ego trajectory with interaction-relevant agent futures. Building on these ego-centric joint mode representations, a causality-aware policy alignment stage further refines the stochastic ego policy using planning-oriented feedback evaluated under the predicted scene context (Sec.~\ref{sec:grpo}).

\subsection{Preliminaries}
\label{sec:prelim}
Our method builds on a query-based end-to-end driving framework~\cite{hipad} that unifies detection, mapping, motion prediction, and planning within a shared task-query architecture. Its unified decoder iteratively refines task queries through temporal attention, agent-agent attention, and interaction attention that couples planning and ego queries with detection and map queries. The resulting marginal agent embeddings encode actor-specific motion cues. We denote the current marginal embedding of agent \(i\) by \(\mathbf{x}^{a}_{i}\in\mathbb{R}^{C}\), and the current ego planning embedding by \(\mathbf{x}^{e}\in\mathbb{R}^{C}\), omitting refinement-layer indices for readability. Throughout this section, \(N\) denotes the number of surrounding agents and \(C\) the feature dimension. The backbone also predicts ego trajectories at multiple granularities: we use the temporally sampled trajectory \(\tau^{e}_{\mathrm{tp}}\) for ego-centric mode assignment and the spatial trajectory \(\tau^{e}_{\mathrm{sp}}\) as a geometric cue for selecting interaction-relevant agents. However, because these marginal embeddings are optimized primarily for actor-wise prediction, they do not explicitly represent shared scene modes conditioned on ego motion. This limitation motivates the joint-causal modeling module introduced next.

\subsection{Ego-Centric Joint-Causal Modeling}
\label{sec:joint_prediction}

CaAD augments the marginal motion prediction branch with a ego-centric joint-causal modeling to improve scene-level consistency between ego planning and surrounding-agent forecasting. The module introduces \(M\) ego-centric scene modes, where each mode \(m\) predicts an ego trajectory \(\hat{\tau}^{e}_{m}\) and surrounding-agent trajectories \(\{\hat{\tau}^{a}_{i,m}\}_{i=1}^{N}\) under the same scene hypothesis. To avoid unnecessary coupling, we apply joint supervision only to agents that are relevant to the ego maneuver, while leaving the remaining agents to the marginal prediction branch.

\paragraph{Marginal--joint embedding interaction.}
Building on the decoded ego and agent embeddings defined in Sec.~\ref{sec:prelim}, we introduce joint-mode embeddings to instantiate ego-centric scene modes. For surrounding agent \(i\), let \(\mathbf{z}^{a}_{i,m}\in\mathbb{R}^{C}\) denote the agent joint-mode embedding associated with the \(m\)-th scene mode, and let \(\mathbf{z}^{e}_{m}\in\mathbb{R}^{C}\) denote the ego joint-mode embedding for the same \(m\)-th mode. We then form ego and agent mode stacks by concatenating each decoded embedding with its \(M\) joint-mode embeddings, namely \(\mathbf{U}^{e}=[\mathbf{x}^{e};\mathbf{z}^{e}_{1};\ldots;\mathbf{z}^{e}_{M}]\) and \(\mathbf{U}^{a}_{i}=[\mathbf{x}^{a}_{i};\mathbf{z}^{a}_{i,1};\ldots;\mathbf{z}^{a}_{i,M}]\), where \(\mathbf{U}^{e}, \mathbf{U}^{a}_{i}\in\mathbb{R}^{(1+M)\times C}\). As illustrated in Fig.~\ref{fig:main}(c), the Agent-Mode Attention module views each stack as a compact mode-wise token sequence, consisting of the decoded embedding followed by \(M\) joint-mode embeddings, and applies attention~\cite{transformer} along the mode dimension:
\begin{equation}
    \bar{\mathbf{U}}^{e}
    =
    \mathrm{Attn}_{\mathrm{AM}}
    \left(
        \mathbf{U}^{e},
        \mathbf{U}^{e},
        \mathbf{U}^{e}
    \right),
    \qquad
    \bar{\mathbf{U}}^{a}_{i}
    =
    \mathrm{Attn}_{\mathrm{AM}}
    \left(
        \mathbf{U}^{a}_{i},
        \mathbf{U}^{a}_{i},
        \mathbf{U}^{a}_{i}
    \right).
\end{equation}
In implementation, we collect the ego and all surrounding-agent stacks into a batch of per-entity mode sequences of shape \((1+M)\times C\), and apply Agent-Mode Attention independently to each sequence. Since the marginal embeddings are produced by the unified decoder, they already contain ego, agent, and map context. Agent-Mode Attention then integrates this entity-level context into each corresponding mode embedding, allowing the joint-mode embeddings to attend scene information specific to the ego or agent entities.

The first output token updates the decoded ego or agent embedding, while the remaining tokens become refined joint-mode embeddings. After the final refinement layer, a joint refinement head decodes the \(m\)-th ego mode embedding and the corresponding agent mode embeddings into an ego-centric scene hypothesis \(\mathcal{S}_{m}=(\hat{\tau}^{e}_{m},\hat{\tau}^{a}_{1,m},\ldots,\hat{\tau}^{a}_{N,m})\), where \(\hat{\tau}^{e}_{m}\) denotes the predicted ego trajectory for mode \(m\), and \(\hat{\tau}^{a}_{i,m}\) denotes the predicted trajectory of agent \(i\) under the same mode. Thus, the shared mode index \(m\) defines an ego-centric future scene hypothesis in which the ego trajectory and surrounding-agent futures are organized under a common scene mode.

\begin{figure}[t!]
  \centering
  \includegraphics[width=1\linewidth]{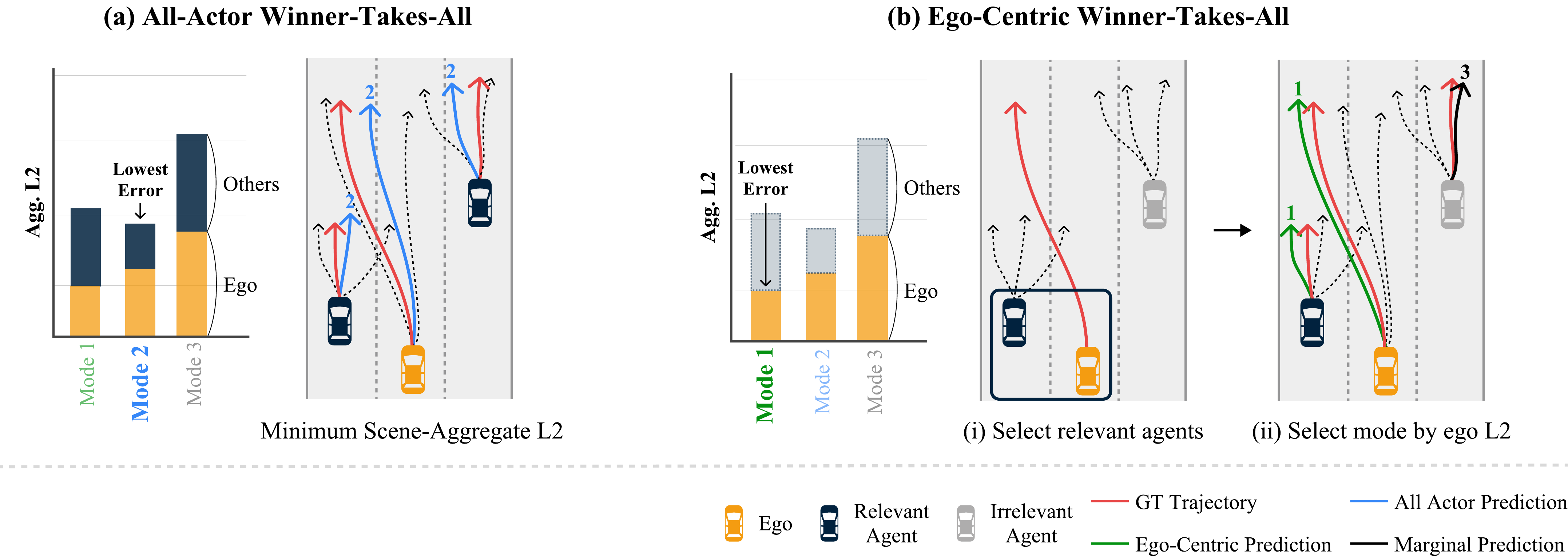}
  \caption{Comparison of joint mode selection strategies. (a) Conventional methods select the joint mode based on the aggregated L2 error over all predicted ego and agent trajectories. 
(b) Ego-centric winner-takes-all selection chooses the joint mode using the ego trajectory L2 error and supervises interaction-relevant agent responses under the selected ego mode.}
  \label{fig:mode_selection}
\vspace{-1.em}
\end{figure}

\paragraph{Interaction-relevant agent selection.}
A key design choice is that we do not force all agents to follow ego-centric joint supervision.
Several agents are far from the ego vehicle and are better handled by standard marginal forecasting.
We therefore select an interaction set $\mathcal{A}_{\mathrm{int}}$ using the ground-truth ego spatial trajectory $\tau^{e}_{\mathrm{sp}}$ and the marginal agent predictions:
\begin{equation}
    \mathcal{A}_{\mathrm{int}}
    =
    \left\{
        i \;\middle|\;
        \exists k \in \mathcal{K}_{i}, \ s.t. \
        \operatorname{collision}\big(\tilde{\tau}^{a}_{i,k}, \tau^{e}_{\mathrm{sp}}\big)=\operatorname{true}
    \right\},
\end{equation}
where $\tilde{\tau}^{a}_{i,k} \in \mathcal{K}_{i}$ is the $k$-th marginal prediction of agent $i$, and $\mathcal{K}_{i}$ denotes the set of marginal multi-modal planning candidates considered for collision checking. We use the spatial trajectory because a temporally sampled ego trajectory does not always expose the intended path clearly, with path geometry often entangled with speed or timing. By contrast, the spatial trajectory directly reflects where the ego intends to move, providing a stable and interpretable cue for interaction relevance. This is also consistent with the sparsity of real-world interactions: only a subset of surrounding agents should meaningfully respond to changes in ego intent, and restricting supervision to this subset avoids over-conditioning irrelevant actors.

\paragraph{Ego-centric mode assignment.}
A conventional all-actor winner-takes-all rule selects the aggregate scene mode, e.g., the one with the lowest summed L2 error for all agent trajectories~\cite{scenetransformer,wayformer,wagner2024jointmotion}. In end-to-end driving, however, we organize the joint representation around the ego trajectory to better align it with the ego planning objective. As shown in Fig.~\ref{fig:mode_selection}, CaAD accordingly selects the joint mode based on the ego trajectory and applies the selected mode only to interaction-relevant agents.
Formally, we define the selected joint scene mode as:
\begin{equation}
    m^{\star}
    =
    \arg\min_{m}
        d(\hat{\tau}^{e}_{m},\tau^{e}_{\mathrm{tp}}),
\end{equation}
where \(d(\cdot,\cdot)\) is a masked trajectory distance.
Agents in \(\mathcal{A}_{\mathrm{int}}\) are then supervised at the same ego-selected mode \(m^{\star}\), while non-selected agents keep marginal supervision with their own best marginal mode \(k^{\star}_{i}=\arg\min_{k}d(\tilde{\tau}^{a}_{i,k},\tau^{a}_{i,\mathrm{gt}})\).
This design focuses scene-mode learning on reciprocal dependencies around ego motion and avoids assigning every actor in the scene to the same joint mode. The joint mode classification loss is formulated as: 
\begin{equation}
    \mathcal{L}_{\mathrm{joint\_cls}}
    =
    \mathcal{L}^{e}_{\mathrm{cls}}(m^{\star})
    +
    \sum_{i\in\mathcal{A}_{\mathrm{int}}}
    \mathcal{L}^{a}_{\mathrm{cls},i}(m^{\star})
    +
    \sum_{i\notin\mathcal{A}_{\mathrm{int}}}
    \mathcal{L}^{a}_{\mathrm{cls},i}(k^{\star}_{i}),
\end{equation}
where \(\mathcal{L}_{\mathrm{cls}}\) denotes the focal loss~\cite{focalloss} for classification.

\subsection{Causality-Aware Policy Alignment}
\label{sec:grpo}

The key novelty of this stage is that policy refinement operates on the ego-centric joint-mode representations learned in Sec.~\ref{sec:joint_prediction}, rather than on isolated marginal or ego-only planning embeddings. Each sampled ego rollout is therefore evaluated together with the corresponding surrounding-agent futures under the same joint mode, grounding the update in a coupled future hypothesis. To improve closed-loop driving quality beyond supervised imitation, we implement this causality-aware policy alignment with a post-training GRPO objective~\cite{shao2024deepseekmath,guo2025deepseekr1}, which is shown in Fig.~\ref{fig:main}(b). Concretely, the ego trajectory head is parameterized as a Gaussian policy. For each joint mode \(m\), we sample \(G\) ego rollouts and score the \(g\)-th rollout with a planning-oriented reward \(r_{m,g}\), using a PDMS-style~\cite{navsim} penalty-plus-quality decomposition adapted for training, as detailed in Appendix~\ref{app:bench2drive_reward}. This aligns the ego policy using complete scene outcomes and encourages safer, more coherent future interactions.

\paragraph{Gaussian trajectory policy.}
Classic E2E methods typically formulate future trajectory prediction as a deterministic regression task.
To incorporate reinforcement learning, however, the model must parameterize a trajectory distribution instead.
We therefore interpret the original regression output as the predictive mean \(\mu\) and additionally predict the standard deviation \(\sigma\).
The supervised joint regression term is then written as a Gaussian negative log-likelihood:
\begin{equation}
\begin{aligned}
    \mathcal{L}_{\mathrm{joint\_reg}}
    &=
    \mathcal{L}^{e}_{\mathrm{reg}}(m^{\star})
    +
    \sum_{i\in\mathcal{A}_{\mathrm{int}}}
    \mathcal{L}^{a}_{\mathrm{reg},i}(m^{\star})
    +
    \sum_{i\notin\mathcal{A}_{\mathrm{int}}}
    \mathcal{L}^{a}_{\mathrm{reg},i}(k^{\star}_{i}), \\
    \mathcal{L}_{\mathrm{reg}}(\tilde{m})
    &=
    \frac{1}{2}
    \left(
        \frac{\bigl(\mu_{\tilde{m}} - \tau_{\mathrm{gt}}\bigr)^2}
        {\sigma_{\tilde{m}}^2}
        +
        2 \log \sigma_{\tilde{m}}
        +
        \log 2\pi
    \right),
\end{aligned}
\end{equation}
where \(\tau_{\mathrm{gt}}\) denotes the corresponding ground-truth trajectory target. This converts the original trajectory regression head into a stochastic policy head during training; during inference, we use only the predicted mean trajectory \(\mu\).

\paragraph{Policy alignment.}
We first sample ego-only rollouts for each mode, represented as \(\bar{\tau}_{m} \in \mathbb{R}^{G \times T_{\mathrm{fut}} \times 2}\), where \(T_{\mathrm{fut}}\) denotes the prediction horizon. We then normalize the rewards \(r_{m,g}\) within each mode group and define the truncated advantage following~\cite{diffusiondrivev2}:
\begin{equation}
A_{m,g}
=
\frac{r_{m,g}-\operatorname{mean}(r_m)}
{\operatorname{std}(r_m)},
\qquad
\tilde{A}_{m,g}
=
\begin{cases}
-1, & \text{if collision occurs},\\
\max(0,A_{m,g}), & \text{otherwise}.
\end{cases}
\end{equation}

We then optimize the clipped GRPO objective over sampled ego rollouts:
\begin{equation}
    \mathcal{L}_{\mathrm{GRPO}}
    =
    -
    \mathbb{E}_{m,g}
    \left[
    \min
    \left(
        \left(\frac{\pi_{\theta}}{\pi_{\theta_{\mathrm{old}}}}\right)\tilde{A}_{m,g},
       \operatorname{clip}\!\left(\frac{\pi_{\theta}}{\pi_{\theta_{\mathrm{old}}}}, 1-\epsilon, 1+\epsilon\right)\tilde{A}_{m,g}
    \right)
    \right],
\end{equation}
where \(\pi_{\theta}\) and \(\pi_{\theta_{\mathrm{old}}}\) denote the current and old policies, respectively, and \(\epsilon\) controls the update step size for training stability. Compared with plain reward maximization, the group-relative formulation is useful here because it asks the model to rank multiple candidate ego rollouts under the same scene context and to move probability mass toward the better ones.
This makes the update closely aligned with the goal of interactive planning, where the critical question is often not whether a trajectory is feasible in isolation, but whether it is preferable to other plausible choices under the same scene. Importantly, we apply reinforcement learning only to the ego policy, while keeping surrounding-agent prediction supervised. This avoids noisy per-agent rewards and stabilizes the joint scene predictor. The total objective functions are deferred to Appendix~\ref{app:training_objective}.

\section{Experiments}
\subsection{Experimental Setup}
We evaluate CaAD on two complementary benchmarks: Bench2Drive~\cite{b2d}, which emphasizes closed-loop performance in interaction-intensive scenarios, and NAVSIM~\cite{navsim}, which evaluates planning quality under its official protocol. On Bench2Drive~\cite{b2d}, we report the official closed-loop metrics together with fine-grained driving-ability scores for challenging maneuvers such as merging, overtaking, and emergency braking. For ablation studies, we additionally use Bench2Drive-mini subset (details in Appendix~\ref{app:b2d_mini}). On NAVSIM~\cite{navsim}, we report PDMS along with its component metrics, which provide a more interpretable breakdown of planning behavior in terms of safety, progress, and ride quality. Our model uses a sparse query-based end-to-end planner backbone~\cite{hipad} and follows the standard training and evaluation setup unless otherwise specified. Additional details on dataset splits, metric definitions, preprocessing, training schedules, optimizer settings, hardware configuration, and ablation settings are provided in Appendix~\ref{app:experiment_details}.

\begin{table}[t]
\vspace{-.5em}
\caption{Comparison of methods on Bench2Drive~\cite{b2d} under open-loop and closed-loop metrics. Avg. L2 is averaged over the 2-second prediction horizon at 2 Hz intervals. $^*$ denotes methods that incorporate RL-based policy refinement during training, and $\dagger$  denotes the ground-truth expert. }
\label{tab:b2d_main}
\vspace{-0.25em}
\centering
\scriptsize
\setlength{\tabcolsep}{6pt}
\renewcommand{\arraystretch}{1.1}
\begin{tabular}{l|c|cccc}
\toprule
\multirow{2}{*}{\textbf{Method}}
& \multicolumn{1}{c|}{\textbf{Open-loop Metric}}
& \multicolumn{4}{c}{\textbf{Closed-loop Metric}} \\
\cmidrule(lr){2-2} \cmidrule(lr){3-6}
& {Avg. L2} $\downarrow$
& {Driving Score} $\uparrow$
& {Success Rate (\%)} $\uparrow$
& {Efficiency} $\uparrow$
& {Comfortness} $\uparrow$ \\
\midrule
Think2Drive$^\dagger$~\cite{think2drive} & --   & 91.85 & 85.41 & --     & --    \\
\midrule
UniAD~\cite{uniad}                  & 0.73 & 45.81 & 16.36 & 129.21 & 43.58 \\
VAD~\cite{vad}                           & 0.91 & 42.35 & 15.00 & 157.94 & 46.01 \\
SparseDrive~\cite{sparsedrive}           & 0.87 & 44.54 & 16.71 & 170.21 & 48.63 \\
GenAD~\cite{genad}                       & --   & 44.81 & 15.90 & --     & --    \\
DriveTransformer~\cite{drivetransformer} & \textbf{0.62} & 63.46 & 35.01 & 100.64 & 20.78 \\
Hydra-NeXt~\cite{hydranext}              & 0.92   & 73.86 & 50.00 & 197.76 & 20.68 \\
MomAD~\cite{momad}                       & 0.82 & 47.91 & 18.11 & 174.91 & \textbf{51.20} \\
HiP-AD~\cite{hipad}                      & 0.69 & 86.77 & 69.09 & \textbf{203.12} & 19.36 \\
VADv2~\cite{vadv2}                       & --   & 76.15 & 50.46 & 178.24 & 37.81 \\
SafeDrive~\cite{kim2026safedrive}        & --   & 66.77 & 42.40 & --     & --    \\
\midrule
DIVER$^*$~\cite{diver}                   & 1.11 & 68.90 & 36.75 & 72.34  & 22.34 \\
ReCogDrive$^*$~\cite{recogdrive}         & --   & 71.36 & 45.45 & 138.18 & 17.45 \\
\midrule
\rowcolor{lightgray} \textbf{CaAD (Ours)}$^*$            & 0.68 & \textbf{87.53} & \textbf{71.81} & 172.53 & 34.34 \\
\bottomrule
\end{tabular}
\end{table}
\begin{table}[t]
\caption{Comparison of different methods under ability metrics. $^*$ denotes methods that incorporate RL-based policy refinement during training}
\label{tab:b2d_ability}
\vspace{-0.25em}
\centering
\scriptsize
\setlength{\tabcolsep}{5.5pt}
\renewcommand{\arraystretch}{1.1}
\begin{tabular}{l|ccccc|c}
\toprule
\multirow{2}{*}{\textbf{Method}} 
& \multicolumn{6}{c}{\textbf{Ability (\%)} $\uparrow$} \\
\cmidrule{2-7}
& {{Merging}}
& {{Overtaking}}
& {{Emergency Brake}}
& {{Give Way}}
& {{Traffic Sign}}
& {{Mean}} \\
\midrule
UniAD~\cite{uniad}                     & 14.10 & 17.78 & 21.67 & 10.00 & 14.21 & 15.55 \\
VAD~\cite{vad}                              & 8.11  & 24.44 & 18.64 & 20.00 & 19.15 & 18.07 \\
DriveTransformer~\cite{drivetransformer} & 17.57 & 35.00 & 48.36 & 40.00 & 52.10 & 38.60 \\
Hydra-NeXt~\cite{hydranext}                 & 40.00 & 64.44 & 61.67 & 50.00 & 50.00 & 53.22 \\
HiP-AD~\cite{hipad}                         & 50.00 & 84.44 & \textbf{83.33} & 40.00 & \textbf{72.10} & 65.98 \\
\midrule
DIVER$^*$~\cite{diver}                      & 35.08 & 25.09 & 41.09 & \textbf{50.00} & 59.21 & 42.09 \\
ReCogDrive$^*$~\cite{recogdrive}            & 29.73 & 20.00 & 69.09 & 20.00 & 71.34 & 42.03 \\
\midrule
\rowcolor{lightgray} \textbf{CaAD (Ours)}$^*$ & \textbf{60.00} & \textbf{86.67} & \textbf{83.33} & \textbf{50.00} & 71.34 & \textbf{70.27} \\
\bottomrule
\end{tabular}
\vspace{-1.5em}
\end{table}

\subsection{Comparison with SOTA Methods}
\myparagraph{Evaluation on Bench2Drive.}
As shown in Tab.~\ref{tab:b2d_main}, we compare CaAD with state-of-the-art methods on the closed-loop Bench2Drive benchmark~\cite{b2d}. CaAD achieves state-of-the-art performance, attaining a Driving Score of 87.53 and a Success Rate of 71.81. Notably, its Driving Score is close to that of Think2Drive~\cite{think2drive}, the expert policy used to collect the training data. This result suggests that the proposed joint-causal scene modeling and causality-aware policy alignment improve scene-level decision making, thereby improving overall planning performance. Tab.~\ref{tab:b2d_ability} further shows that the advantage is particularly pronounced for interaction-critical abilities, such as \textit{Merging} and \textit{Overtaking}, where it is essential to reason about how surrounding agents respond to the ego maneuver. The methods marked with $^*$ provide a particularly informative comparison: DIVER~\cite{diver} and ReCogDrive~\cite{recogdrive} incorporate RL-based refinement and outperform several purely imitation-learned methods, yet their improvements remain uneven across abilities. This suggests that policy refinement alone is insufficient when the underlying scene representation remains marginal. Overall, this pattern supports the effectiveness of the proposed joint-causal scene modeling, which provides a causality-aware scene representation that makes subsequent policy alignment substantially more reliable.

\begin{table*}[t]
\caption{Compact ablation on Bench2Drive-mini. Checkmarks indicate temporal/spatial ego cues for relevant-agent selection, joint prediction on selected agents with ego-centric mode assignment, and policy alignment on marginal or joint scene representations. $\dagger$ denotes reproduced under the same 6-epoch ablation setting.}
\label{tab:abl_components}
\vspace{0.5em}
\centering
\tiny
\setlength{\tabcolsep}{3.6pt}
\renewcommand{\arraystretch}{1.1}
\providecommand{\up}[1]{{\tiny(\textcolor{red}{#1$\uparrow$})}}
\providecommand{\dn}[1]{{\tiny(\textcolor{blue}{#1$\downarrow$})}}
{%
\begin{tabular}{l|cc|cc|cc|cc|c}
\toprule
\multirow{2}{*}{\textbf{Model}}
& \multicolumn{2}{c|}{\textbf{Agent Selection}}
& \multicolumn{2}{c|}{\textbf{Joint Prediction}}
& \multicolumn{2}{c|}{\textbf{Policy Align}}
& \multicolumn{2}{c|}{\textbf{Closed-loop}}
& \textbf{Ability} \\
\cmidrule(lr){2-3}
\cmidrule(lr){4-5}
\cmidrule(lr){6-7}
\cmidrule(lr){8-9}
\cmidrule(lr){10-10}
& {Temp. Cue}
& {Spatial Cue}
& {Selected Agents}
& {Ego-centric}
& {Marginal}
& {Joint}
& {DS} $\uparrow$
& {SR(\%)} $\uparrow$
& Mean $\uparrow$ \\
\midrule
HiP-AD$^\dagger$ (Base)~\cite{hipad}
&  &  &  &  &  & 
& 83.12 & 62.96 & 63.83 \\
\midrule
Model A
&  &  &  &  & \checkmark & 
& 87.43 \up{4.31} & 64.81 \up{1.85} & 66.38 \up{3.45} \\
Model B
& \checkmark &  & \checkmark & \checkmark &  & 
& 86.90 \up{3.78} & 68.52 \up{5.56} & 66.92 \up{3.09} \\
Model C
&  & \checkmark & \checkmark &  &  & 
& 84.76 \up{1.64} & 64.81 \up{1.85} & 56.29 \dn{7.54} \\
Model D
&  & \checkmark & \checkmark & \checkmark &  & 
& 88.61 \up{5.49} & 75.93 \up{12.97} & 72.66 \up{8.83} \\
\rowcolor{lightgray}
\textbf{Model E (Ours)}
&  & \checkmark & \checkmark & \checkmark &  & \checkmark
& \textbf{91.05} \up{7.93} & \textbf{79.63} \up{16.67} & \textbf{76.51} \up{12.68} \\
\bottomrule
\end{tabular}%
}
\vspace{-2.5em}
\end{table*}
\begin{wraptable}{r}{0.5\textwidth}
\vspace{-1.8em}
\caption{Closed-loop planning performance on NAVSIM~\cite{navsim}. The upper block lists non-RL methods, and the lower block lists RL-based methods.}
\label{tab:navsim_main}
\vspace{0.5em}
\centering
\scriptsize
\setlength{\tabcolsep}{4pt}
\renewcommand{\arraystretch}{0.95}
{%
\begin{tabular}{l|cccc|c}
\toprule
\multirow{2}{*}{\textbf{Method}} & \multicolumn{4}{c|}{\textbf{PDMS Components}} & \multirow{2}{*}{\textbf{PDMS}$\uparrow$} \\
\cmidrule(lr){2-5}
& \textbf{NC}$\uparrow$ & \textbf{DAC}$\uparrow$ & \textbf{EP}$\uparrow$ & \textbf{TTC}$\uparrow$ & \\
\midrule
PARA-Drive~\cite{paradrive} & 97.9 & 92.4 & 79.3 & 93.0 & 84.0 \\
LAW~\cite{law} & 96.4 & 95.4 & 81.7 & 88.7 & 84.6 \\
GoalFlow~\cite{goalflow} & 98.3 & 93.8 & 79.8 & 94.3 & 85.7 \\
DiffusionDrive~\cite{diffusiondrive} & 98.2 & 96.2 & 82.2 & 94.7 & 88.1 \\
Hydra-MDP++~\cite{hydramdp++} & 97.6 & 96.0 & 80.4 & 93.1 & 86.6 \\
WOTE~\cite{wote} & 98.4 & 96.6 & 81.7 & 94.5 & 88.0 \\
\midrule
\multicolumn{6}{l}{\textit{RL-based methods}} \\
\midrule
DIVER~\cite{diver} & 98.5 & 96.5 & 82.6 & 94.9 & 88.3 \\
DriveDPO~\cite{shang2025drivedpo} & 98.5 & 98.1 & 84.3 & 94.8 & 90.0 \\
ReCogDrive~\cite{recogdrive} & 97.9 & 97.3 & \textbf{87.3} & 94.9 & 90.8 \\
\midrule
\midrule
\rowcolor{lightgray}
\textbf{CaAD (Ours)} & \textbf{99.4} & \textbf{98.3} & 84.9 & \textbf{95.9} & \textbf{91.1} \\
\bottomrule
\end{tabular}%
}
\vspace{-1.2em}
\end{wraptable}

\myparagraph{Evaluation on NAVSIM.} As shown in Table~\ref{tab:navsim_main}, CaAD also performs strongly on NAVSIM~\cite{navsim}, achieving the best PDMS of 91.1 and the strongest collision-related results. The gains are not confined to the aggregate score: compared with prior RL-used methods~\cite{diver,shang2025drivedpo,recogdrive}, CaAD consistently improves safety-critical components, where reasoning about ego-agent interactions is particularly important. This result suggests that policy alignment is more effective when applied to causality-aware scene embeddings than to marginal representations. In CaAD, policy alignment refines an ego policy that has already been shaped by joint-causal scene modeling, using rewards computed from the ground-truth futures of the jointly considered agents together with map context. As a result, the aligned policy produces safer rollouts with larger collision margins while maintaining strong rule compliance and competitive route progress.

\subsection{Ablation Studies}
\myparagraph{Effectiveness of joint-mode design and causality-aware policy alignment.}  Tab.~\ref{tab:abl_components} shows that joint-causal scene modeling is already highly effective before policy alignment. Compared to the reproduced baseline, Model D, which incorporates spatial agent selection, joint prediction of the selected agents, and ego-centric assignment, improves Driving Score, Success Rate, and Ability mean by an average of 5.49, 12.97, and 8.83 points, respectively. This suggests that interactive planning benefits from organizing futures of relevant agents under the ego-selected scene mode, rather than relying solely on marginal predictions. Building on this stronger representation, causality-aware policy alignment (Model E) yields the best overall performance, further improving over Model D by 2.44 Driving Score, 3.70 Success Rate, and 3.85 Ability mean points, and over the reproduced baseline by 7.93, 16.67, and 12.68 points, respectively. 

Importantly, these gains do not arise from policy alignment alone. Although Model A performs policy alignment directly on marginal embeddings, it improves over the reproduced baseline by only 4.31 Driving Score, 1.85 Success Rate, and 3.45 Ability mean points. In contrast, Model D already surpasses Model A by 1.18, 11.12, and 6.28 points without policy alignment, and Model E further exceeds Model A by 3.62, 14.82, and 10.13 points. Together, these comparisons show that joint-causal scene modeling provides a stronger planning representation than marginal policy alignment alone, serving as a prerequisite for effective policy alignment in interactive driving.

\myparagraph{Impact of ego trajectory representation for agent selection.} Tab.~\ref{tab:abl_components} compares temporal and spatial ego cues for selecting interaction-relevant agents. Under the same joint-prediction and ego-centric assignment setting, replacing the temporal cue in Model B with the spatial cue in Model D improves Driving Score, Success Rate, and Ability mean by 1.71, 7.41, and 5.74 points, respectively. This suggests that the spatial cue provides a clearer signal of the ego maneuver for relevant-agent selection, whereas the temporal cue can be entangled with speed or timing and lead to weaker supervision.

\myparagraph{Importance of Ego-Centric Assignment.} Tab.~\ref{tab:abl_components} also isolates role of ego-centric assignment is even more direct. When spatial selection and joint prediction are kept fixed, removing ego-centric assignment from Model D to obtain Model C lowers Driving Score, Success Rate, and Ability mean by 3.85, 11.12, and 16.37 points, respectively. This substantial degradation shows that identifying relevant agents is not sufficient; their futures must also be organized under the ego-selected scene mode. These results validate ego-centric assignment as a key component for effective interaction-sparse supervision.

\subsection{Qualitative Results}
\begin{figure}[t!]
  \centering
  \includegraphics[width=1\linewidth]{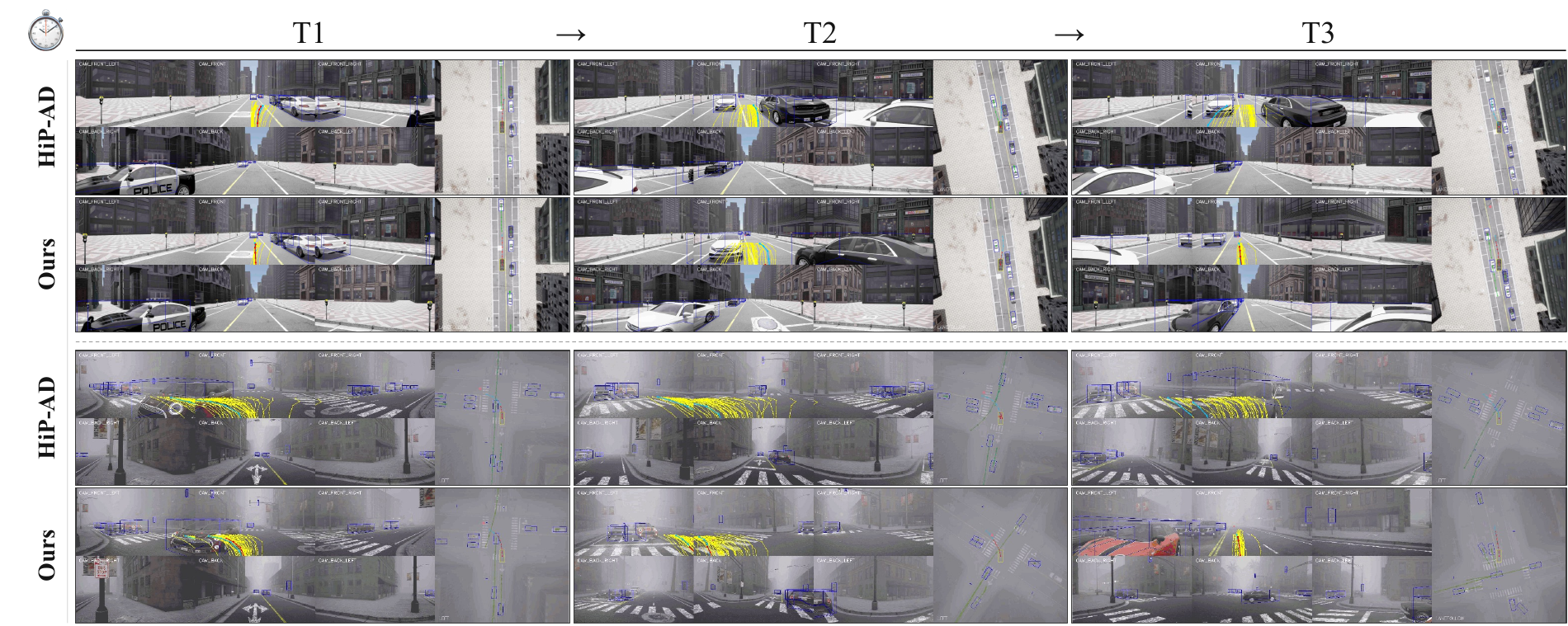}
  \vspace{-1em}
  \caption{Qualitative examples of interaction-aware planning. We visualize consecutive frames in two challenging interactive scenarios, where the \textcolor{cyan}{sky-blue trajectory} denotes the spatial trajectory and the \textcolor{red}{red trajectory} denotes the temporal trajectory.
}
  \label{fig:qualitative}
\end{figure}

Fig.~\ref{fig:qualitative} shows qualitative results of CaAD in challenging interactive scenarios. (a) CaAD reasons about surrounding vehicles and proceeds through the gap between agents, avoiding the deadlock that may occur when agent responses are not modeled jointly with the ego plan. (b) CaAD first waits for the interacting agent to pass before proceeding, whereas a planner without joint interaction modeling may enter prematurely, stop in front of the agent, and fail to make progress. These results show that ego-conditioned joint prediction enables CaAD to produce more coherent and socially compliant closed-loop behaviors in dense multi-agent traffic. Additional qualitative results on NAVSIM~\cite{navsim}, together with further qualitative examples for the ablation study, are provided in the Appendix~\ref{app:addqual}.

\section{Conclusion}
We propose CaAD, a Causality-aware end-to-end Autonomous Driving framework that combines joint-causal scene modeling with causality-aware policy alignment for planning-oriented scene modeling. By learning ego-conditioned joint scene representations, CaAD achieves state-of-the-art performance on Bench2Drive and NAVSIM while improving interaction-consistent planning. We hope CaAD encourages a new direction for end-to-end driving, where future planning is grounded not only in ego trajectory prediction but also in causality-aware modeling of interactive scene dynamics. Handling long-tail social behaviors and validating transfer to more diverse real-world settings remain important directions for future work.

\bibliographystyle{plain}
\bibliography{bib}

\clearpage


\appendix

\appendix
\makeatletter
\newcommand{\makeappendixtitle}{
  \clearpage
  \begingroup
    \thispagestyle{empty}
    \vbox{
      \hsize\textwidth
      \linewidth\hsize
      \vskip 0.1in
      \@toptitlebar
      \centering
      {\LARGE\bf Appendix\par}
      \@bottomtitlebar
    }
  \endgroup
}
\makeatother
\makeappendixtitle

\renewcommand{\thefigure}{A\arabic{figure}}
\setcounter{figure}{0}
\renewcommand{\thetable}{A\arabic{table}}
\setcounter{table}{0}
\renewcommand{\thesection}{\Alph{section}}
\setcounter{section}{0}

\section{Additional Related Work}
\label{app:related_work_details}

\subsection{End-to-End Planning for Autonomous Driving}
End-to-end autonomous driving learns planning-oriented representations directly from sensor observations, offering an alternative to modular pipelines~\cite{chen2024end,transfuser,chitta2021neat,uniad,vad,sparsedrive, transfuser++, hipad, genad}. Early methods such as TransFuser~\cite{transfuser} and NEAT~\cite{chitta2021neat} predict planning-relevant waypoints from fused sensor features or BEV-based attention-field representations. More recent approaches incorporate stronger intermediate structures. UniAD~\cite{uniad} unifies perception, prediction, and planning in a planning-oriented framework, while VAD~\cite{vad} and subsequent works~\cite{sparsedrive,diffusiondrive,hipad,genad} improve vectorized scene modeling, efficiency, multimodal ego planning, or generative future modeling. To better capture multi-modality and uncertainty, generative planning methods have further emerged, ranging from anchor-based methods~\cite{vadv2, sparsedrive, sparsedrivev2, hipad} to diffusion- or flow-based trajectory generation~\cite{genad, diffusiondrive, goalflow}. Recent studies have also explored world-model-based planning, motivated by its ability to model future scene evolution~\cite{think2drive, raw2drive, world4drive, law, wote, seerdrive, resworld}. Nevertheless, the final objective in much of this literature remains tightly centered on ego planning. These approaches lack a fully explicit joint model of ego-agent future dynamics. Consequently, reciprocal dependencies in highly interactive driving scenarios may be captured only partially or implicitly.

\subsection{Reinforcement Learning for End-to-End Planning}
Although end-to-end driving is dominated by imitation learning (IL), reinforcement learning (RL) provides supervisory signals for objectives beyond behavior cloning, such as safety and human-preference alignment~\cite{chekroun2023gri,jiang2025alphadrive,liang2018cirl,lu2023imitation,li2025learning,shang2025drivedpo, alpamayo, recogdrive, autodrive}. Early methods such as CIRL~\cite{liang2018cirl} and GRI~\cite{chekroun2023gri} combine expert demonstrations with RL for vision-based autonomous driving. More recent work performs post-training refinement of supervised planners or policies, including BC-SAC~\cite{lu2023imitation}, TrajHF~\cite{li2025learning}, and DriveDPO~\cite{shang2025drivedpo}, using RL or preference-based objectives to improve safety, reasoning, or preference alignment. Recent works have further improved reasoning and planning capabilities beyond imitation-centric training~\cite{alpamayo, recogdrive, autodrive, jiang2025alphadrive} for vision-language-action (VLA) models. However, these approaches still optimize primarily the ego policy or planner. RL serves mainly as an auxiliary signal beyond IL, rather than as an explicit objective for jointly cooperative multi-agent trajectory generation. Instead, our proposed CaAD employs a joint-causal policy alignment stage to fine-tune the driving policy with rewards that promote globally coherent, safe, and socially compliant behavior.

\subsection{Joint Prediction for Multi-Agent Interaction}
In interactive driving, independently predicted marginal futures can be mutually inconsistent, motivating joint models for scene-consistent futures. Scene Transformer~\cite{scenetransformer} introduces a scene-centric transformer for joint, marginal, and conditional prediction, while M2I~\cite{sun2022m2i} improves tractability by factorizing interactions into influencer--reactor pairs. Other interaction-aware approaches include JFP~\cite{luo2023jfp}, which learns consistent multi-agent trajectories; AutoBots~\cite{girgis2022autobots}, which models joint trajectory distributions with latent-variable set transformers; MTR~\cite{shi2022mtr}, which combines intention localization with refinement; MotionLM~\cite{seff2023motionlm}, which autoregressively models discrete motion tokens; and JointMotion~\cite{wagner2024jointmotion}, which pretrains joint predictors via self-supervision. More recently, RetroMotion~\cite{wagner2025retromotion} decomposes forecasting into marginal and pairwise joint predictions, enabling retrocausal refinement and instruction-based modification. Complementary to these forecasting methods, SafeDrive~\cite{kim2026safedrive} adds fine-grained safety reasoning to end-to-end driving by estimating agent-specific collision risks and drivable-region compliance over future timesteps. Overall, these works mainly address interaction-aware prediction and safety-aware ego planning, rather than explicitly optimizing a globally coordinated multi-agent plan.

\section{Additional Method Details}
\label{app:method_details}

\subsection{Reward Function}
\label{app:bench2drive_reward}

For RL stage on Bench2Drive~\cite{b2d}, each sampled ego rollout is evaluated together with the jointly predicted futures of surrounding agents under the same scene mode. We follow the penalty-plus-quality decomposition of PDMS used in NAVSIM~\cite{navsim}, but adapt several subscores to make the reward more suitable for scene-level GRPO~\cite{shao2024deepseekmath}. In particular, the original discrete PDMS subscores can assign identical rewards to many rollouts sampled within the same scene group. Such ties reduce the informativeness of the relative ranking signal used by GRPO. We therefore preserve the safety-oriented structure of PDMS while making selected subscores continuous.

\begin{equation}
\label{eq:bench2drive_reward}
r
=
\underbrace{\left(\prod_{p \in \{NC,DAC,DD\}} \text{score}_p \right)}_{\text{safety/rule penalties}}
\times
\underbrace{\left(\frac{\sum_{w \in \{EP,TTC,C\}} \text{weight}_w \times \text{score}_w}{\sum_{w \in \{EP,TTC,C\}} \text{weight}_w}\right)}_{\text{driving-quality term}} .
\end{equation}

Here, NC denotes no at-fault collision, DAC drivable-area compliance, DD driving-direction compliance, EP ego progress, TTC time-to-collision margin, and C comfort. We use weights $5.0$, $5.0$, and $2.0$ for EP, TTC, and C, respectively.

\paragraph{Continuous subscore modifications.}
We modify the following subscores. The no-collision subscore remains discrete because at-fault collisions are safety-critical. These changes produce denser rewards and reduce ties among rollouts sampled in the same scene group, making the reward better matched to GRPO's relative comparison objective.

\textbf{Driving-direction compliance.}
Let $d_{\mathrm{opp}}$ denote the maximum accumulated opposite-direction progress. Instead of using the original three-level score, we linearly interpolate between the compliance threshold $2.0$\,m and violation threshold $6.0$\,m:
\[
\text{score}_{DD} =
\begin{cases}
1.0, & d_{\mathrm{opp}} \le 2.0, \\
1 - \frac{d_{\mathrm{opp}} - 2.0}{6.0 - 2.0}, & 2.0 < d_{\mathrm{opp}} < 6.0, \\
0.0, & d_{\mathrm{opp}} \ge 6.0.
\end{cases}
\]

\textbf{Drivable-area compliance.}
Instead of assigning zero reward whenever any off-road timestep occurs, we use the fraction of timesteps that remain within the drivable area:
\[
\text{score}_{DAC}
=
1 -
\frac{1}{T}
\sum_{t=1}^{T}
\mathbbm{1}\{\text{ego is outside the drivable area at } t\}.
\]

\textbf{Time-to-collision.}
Instead of a binary TTC score, we score the rollout by the first TTC-infraction timestep:
\[
\text{score}_{TTC}
=
\frac{\min(t_{\mathrm{TTC}}, T)}{T},
\]
where $t_{\mathrm{TTC}}$ is the first TTC-infraction timestep and $T$ is the rollout horizon. If no infraction occurs, the score is $1$.

\textbf{Comfort.}
Instead of a hard conjunction over comfort thresholds, each comfort metric receives a continuous violation score:
\[
s_k = \exp\left(-\frac{\Delta_k}{\alpha_k}\right),
\qquad
\text{score}_{C} = \min_k s_k,
\]
where $\Delta_k$ is the violation magnitude for comfort metric $k$, and $\alpha_k$ is a scale factor based on the corresponding threshold.

\subsection{Training Objective}
\label{app:training_objective}
For completeness, we collect the full optimization objective here. Let \(\mathcal{L}_{\mathrm{e2e}}\) denote the original supervised objective, including detection, mapping, planning, and marginal motion prediction~\cite{hipad}. The final objective is
\begin{equation}
    \mathcal{L}
    =
    \mathcal{L}_{\mathrm{e2e}}
    +
    \lambda_{\mathrm{joint}}(\mathcal{L}_{\mathrm{joint\_reg}} + \mathcal{L}_{\mathrm{joint\_cls}})
    +
    \lambda_{\mathrm{rl}}\mathcal{L}_{\mathrm{GRPO}},
\end{equation}
where \(\lambda_{\mathrm{joint}}\) and \(\lambda_{\mathrm{rl}}\) are weighting coefficients that control the contributions of the joint prediction loss and the reinforcement learning objective, respectively. This design preserves reliable scene-conditioned prediction under supervision, while allowing RL to refine only the preference over ego actions. As a result, supervised ego-conditioned joint prediction captures how the scene is organized around ego intent, whereas scene-aware GRPO improves closed-loop action selection under the predicted scene.

\section{Additional Experimental Details}
\label{app:experiment_details}

\subsection{Dataset and Metrics}
\label{app:dataset_metrics}
\myparagraph{Bench2Drive and Metrics.} Our primary evaluation uses Bench2Drive~\cite{b2d}, a challenging closed-loop benchmark designed to stress interactive driving. It provides 950 clips for training and 50 clips for open-loop validation, while closed-loop performance is measured on 220 predefined routes. We report the official metrics, including Driving Score (DS), Success Rate (SR), Efficiency, and Comfortness, and further break performance down into interaction-centric driving abilities such as Merging, Overtaking, Emergency Braking, Give Way, and Traffic Sign compliance. Ablation studies are conducted on the subsampled Bench2Drive benchmark~\cite{b2d}, which is representative of the full dataset.

\myparagraph{NAVSIM and Metrics.} We additionally evaluate on NAVSIM~\cite{navsim}, a planning-oriented benchmark based on data-driven non-reactive simulation. Under the official NAVSIM v1 protocol, each predicted ego trajectory is rolled out for 4 seconds, while background actors follow their logged future trajectories. This setup measures planning quality in a standardized setting that isolates the effect of the ego policy. The primary NAVSIM metric is the Predictive Driver Model Score (PDMS), which summarizes planning performance by combining safety-critical penalties with a weighted quality term~\cite{navsim}. We also report its component metrics: no at-fault collisions (NC), drivable area compliance (DAC), ego progress (EP), and time-to-collision within bound (TTC). NC and DAC capture safety and rule compliance, whereas EP, and TTC reflect route progress, collision margin, and ride smoothness. Reporting both PDMS and its components makes the evaluation easier to interpret, since improvements may arise from different aspects of planning behavior. Since the NAVSIM planning metrics are also used to construct the reward in our scene-aware RL stage.

\subsection{Bench2Drive-mini for Ablation Studies}
\label{app:b2d_mini}
For closed-loop ablation experiments, we use the Bench2Drive-mini split~\cite{hipad}. Evaluating every ablation on all 220 routes is computationally expensive, so we utilized the 54-route subset, corresponding to around 25\% of the full benchmark. This subset includes one representative route from each of the 43 scenarios, together with 11 additional randomly selected routes. Therefore, it preserves full scenario coverage while serving as a computationally efficient proxy for the full benchmark. We use exactly the same Route IDs listed in Tab.~\ref{tab:b2d_mini_routes}.

\begin{table*}[h]
\vspace{-1.em}
\centering
\small
\caption{Bench2Drive-mini~\cite{b2d} Route IDs used for closed-loop ablations.}
\vspace{0.5em}
\label{tab:b2d_mini_routes}
\begin{tabular}{p{0.16\textwidth} p{0.78\textwidth}}
\toprule
\textbf{Split} & \textbf{Route IDs} \\
\midrule
43 scenario routes & 1711, 1773, 1790, 1825, 1852, 1956, 2050, 2082, 2084, 2115, 2127, 2144, 2164, 2201, 2204, 2273, 2373, 2390, 2416, 2509, 2534, 2664, 2709, 2790, 3086, 3248, 3364, 3436, 3464, 3540, 3561, 3936, 14194, 14842, 17563, 17752, 23658, 23695, 23771, 23901, 26458, 28087, 28099 \\
\midrule
11 random routes & 1792, 2086, 2129, 2283, 2539, 2668, 26406, 26956, 27494, 27532, 28154 \\
\bottomrule
\end{tabular}
\end{table*}

\subsection{Implementation Details}
\label{app:implementation_details}
\myparagraph{Implementation for Bench2Drive.} Our implementation uses a sparse query-based planner with multi-granularity ego queries, following the released baseline configuration~\cite{hipad} unless otherwise specified. Input images are resized to \(640 \times 352\), and target points and high-level commands are embedded into the planner. Ego status is predicted during both training and inference. In addition, we sample 2m trajectories for \(\tau^e_{\mathrm{sp}}\). We use the predicted temporal and spatial trajectories only for longitudinal and lateral control, respectively, while ego-centric joint prediction and SRL are used solely as training-time supervision.

Training is conducted in three stages. We first train the detection module for 12 epochs, followed by the planning module with joint prediction for another 12 epochs. We then further optimize the model with reinforcement learning for 6 epochs. All experiments are conducted on 8 NVIDIA A6000 GPUs with a total batch size of 48. The initial learning rate is set to \(2 \times 10^{-4}\) for the first two stages and \(2 \times 10^{-5}\) for the reinforcement learning stage. We use the AdamW optimizer~\cite{adamw} with a weight decay of 0.01. For the ablation study, we train all models under the same setting, with 6 epochs for stage two and 3 epochs for stage three.

\myparagraph{Implementation for NAVSIM.} For NAVSIM, we follow the implementation of DiffusionDrive~\cite{diffusiondrive}. We use ResNet-34~\cite{resnet} as the image encoder and additionally provide rasterized BEV LiDAR as input. Three front-view camera images are downsampled and concatenated into an input resolution of \(1024 \times 256\). The model is trained from scratch on the navtrain split for 80 epochs using the AdamW optimizer~\cite{adamw} with a learning rate of \(6 \times 10^{-4}\), and is then further optimized for 20 epochs with an additional PDMS-based causality-aware policy alignment stage.

\section{Additional Experiments}
\label{app:experiment_additional}

\subsection{Additional Quantitative Results}
\myparagraph{Ego-centric vs. All-agent Policy Alignment.}
We further ablate the scope of reinforcement learning by comparing ego-centric joint policy alignment with all-agent policy alignment, where rollout sampling and policy optimization are applied not only to the ego vehicle but also to each surrounding agent. Tab.~\ref{tab:ego_scene_rl} shows that all-agent policy alignment slightly improves Driving Score and Ability mean, but reduces Success Rate. In contrast, ego-centric causality-aware policy alignment achieves the best overall performance. This indicates that applying alignment to all agents is not necessarily beneficial, as per-agent sampling can introduce noisy credit assignment and destabilize supervised surrounding-agent forecasting. Ego-centric alignment instead preserves a reliable joint scene predictor and focuses policy optimization on the ego decisions that directly determine closed-loop performance.

\begin{table}[t]
\caption{Ablation on Bench2Drive~\cite{b2d} full routes set. $\dagger$ denotes reproduced results.}
\label{tab:ego_scene_rl}
\vspace{-0.25em}
\centering
\scriptsize
\setlength{\tabcolsep}{7pt}
\renewcommand{\arraystretch}{1.02}

\providecommand{\up}[1]{{\tiny(\textcolor{red}{#1$\uparrow$})}}
\providecommand{\dn}[1]{{\tiny(\textcolor{blue}{#1$\downarrow$})}}

\begin{tabular}{l|cccc|c}
\toprule
\multirow{2}{*}{\textbf{Model}} 
& \multicolumn{4}{c|}{\textbf{Closed-loop}} 
& \textbf{Ability} \\
\cmidrule(lr){2-5} \cmidrule(lr){6-6}
& {DS} $\uparrow$ & {SR (\%)} $\uparrow$ & {Efficiency} $\uparrow$ & {Comfortness} $\uparrow$ & {Mean} $\uparrow$ \\
\midrule
A. HiP-AD$^\dagger$ (Base)~\cite{hipad} & 86.38 & 66.82 & 200.11 & 16.89 & 63.33 \\
\midrule
B. All-agent Policy Alignment& 86.90 \up{0.52} & 65.45 \dn{1.37} & 203.39 \up{3.28} & 14.73 \dn{2.16} & 65.36 \up{2.03} \\
C. Ego-centric Policy Alignment (Ours)& 87.53 \up{1.15} & 71.81 \up{4.99} & 172.53 \dn{27.58} & 34.34 \up{17.45} & 70.27 \up{6.94} \\
\bottomrule
\end{tabular}
\vspace{-1.5em}
\end{table}
\begin{wraptable}{R}{0.56\columnwidth}
\vspace{-2em}
\caption{Open-loop evaluation on nuScenes~\cite{nuscenes} \textit{val} set. \(\dagger\) indicates reproduced results.}
\label{tab:nus}
\centering
\scriptsize
\renewcommand{\arraystretch}{1.03}
\setlength{\tabcolsep}{3pt}
{%
\begin{tabular}{l|ccc>{\columncolor{gray!20}}c|ccc>{\columncolor{gray!20}}c}
\toprule
\multirow{2}{*}{\textbf{Method}} 
& \multicolumn{4}{c|}{\textbf{L2 (m)} $\downarrow$}
& \multicolumn{4}{c}{\textbf{Collision (\%)} $\downarrow$} \\
\cmidrule(lr){2-5} \cmidrule(lr){6-9}
& \textbf{1s} & \textbf{2s} & \textbf{3s} & \textbf{Avg.}
& \textbf{1s} & \textbf{2s} & \textbf{3s} & \textbf{Avg.} \\
\midrule
VAD-Base~\cite{vad} & 0.41 & 0.70 & 1.05 & 0.72 & 0.03 & 0.19 & 0.43 & 0.21 \\
GenAD~\cite{genad} & 0.28 & 0.49 & 0.78 & 0.52 & 0.08 & 0.14 & 0.34 & 0.19 \\
SparseDrive~\cite{sparsedrive} & 0.29 & 0.58 & 0.96 & 0.61 & 0.01 & 0.05 & 0.18 & 0.08 \\
DriveTransformer~\cite{drivetransformer} & 0.16 & 0.30 & 0.55 & 0.33 & 0.01 & 0.06 & 0.15 & 0.07 \\
MomAD~\cite{momad} & 0.31 & 0.57 & 0.91 & 0.60 & 0.01 & 0.05 & 0.22 & 0.09 \\
HiP-AD~\cite{hipad} & 0.28 & 0.53 & 0.87 & 0.56 & 0.01 & 0.05 & 0.15 & 0.07 \\
\midrule
HiP-AD$^\dagger$~\cite{hipad} & 0.31 & 0.63 & 1.05 & 0.67 & 0.00 & 0.03 & 0.14 & 0.06 \\
CaAD (Ours) & 0.28 & 0.56 & 0.94 & 0.59 & 0.02 & 0.02 & 0.19 & 0.08 \\
\bottomrule
\end{tabular}%
}
\vspace{-1em}
\end{wraptable}

\myparagraph{Evaluation on the nuScenes Open-loop Setting.}
We additionally evaluate CaAD in the open-loop setting on nuScenes~\cite{nuscenes}. Compared with the reproduced baseline~\cite{hipad}, CaAD achieves a lower L2 error but a slightly higher collision rate. This implies that CaAD optimizes a closed-loop scene-level interaction, suggesting a performance mismatch can occur between closed-loop and open-loop evaluation~\cite{pdmhybrid}. Despite this, CaAD shows stronger closed-loop performance, suggesting improved interaction-aware planning beyond open-loop trajectory matching.

\clearpage
\begin{figure}[t!]
  \centering
  \includegraphics[width=1\linewidth]{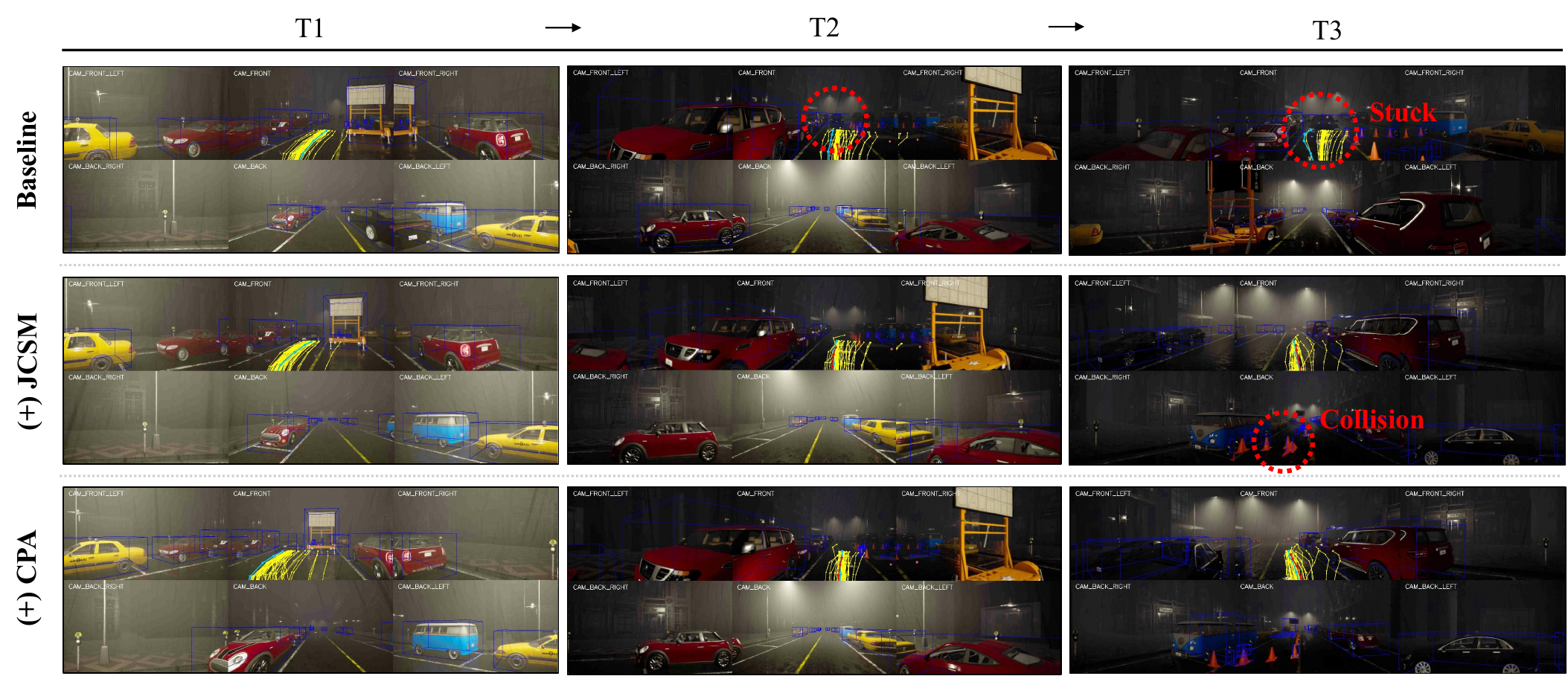}
  \vspace{-1em}
  \caption{Qualitative examples of interaction-aware planning under module ablations.
JCSM and CPA denote joint-causal scene modeling and causality-aware policy alignment, respectively. 
We visualize consecutive frames in challenging interactive scenarios, where the \textcolor{cyan}{sky-blue trajectory} denotes the spatial trajectory and the \textcolor{red}{red trajectory} denotes the temporal trajectory.
}
  \label{fig:apdx_qualitative}
\end{figure}

\begin{figure}[t!]
  \centering
  \includegraphics[width=1\linewidth]{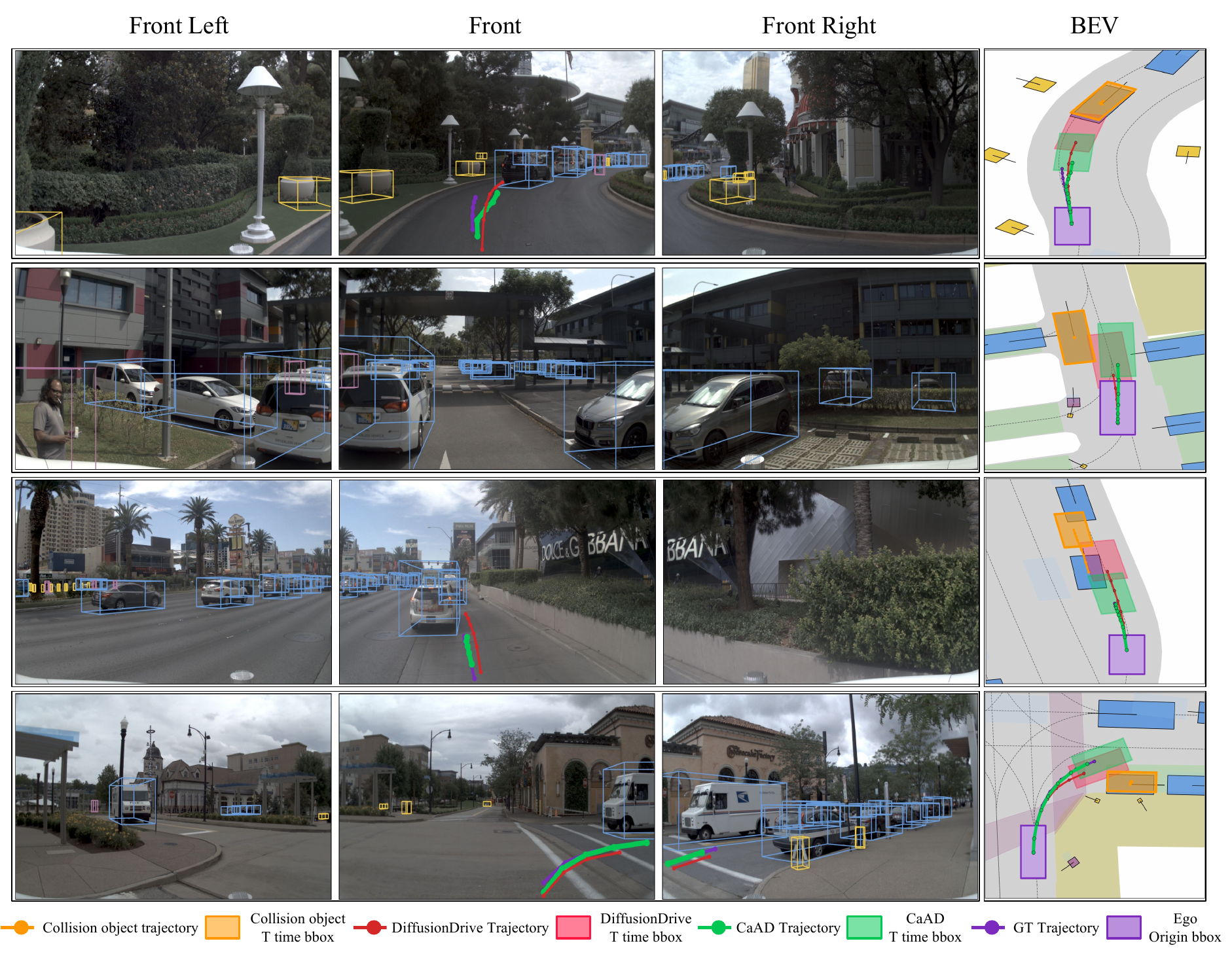}
  \vspace{-1em}
  \caption{Qualitative NAVSIM~\cite{navsim} collision cases. Each row shows three camera views and BEV. 
\textcolor{caribbeangreen}{Green}, \textcolor{red}{red}, and \textcolor{amethyst}{purple} trajectories denote CaAD, DiffusionDrive~\cite{diffusiondrive}, and ground truth, respectively. 
In BEV, all boxes except the \textcolor{amethyst}{purple} ego box are drawn at the same future time step \(T\): 
\textcolor{orange}{orange} denotes the collision-related object, 
\textcolor{red}{red} denotes the DiffusionDrive-predicted ego pose, and 
\textcolor{caribbeangreen}{green} denotes the CaAD-predicted ego pose. 
The \textcolor{amethyst}{purple} box denotes the current ego pose at \(t=0\).
}
  \label{fig:apdx_navsim}
\end{figure}

\subsection{Additional Qualitative Results}
\label{app:addqual}
\myparagraph{Qualitative Comparison of Components.}
Fig.~\ref{fig:apdx_qualitative} compares qualitative behaviors under different module configurations. The baseline~\cite{hipad} tends to exploit locally available free space without sufficiently reasoning about surrounding-agent interactions, which can cause the ego vehicle to stop and become stuck in front of another vehicle. Ego-centric joint-causal modeling improves interaction reasoning by organizing relevant agent futures around ego motion, but may still suffer from failures such as colliding with a traffic cone. With Causal-aware policy alignment, the ego policy is further aligned using planning-oriented feedback, producing safer and more coherent interaction-aware behavior.

\myparagraph{Qualitative Results for NAVSIM.}
Fig.~\ref{fig:apdx_navsim} shows collision cases on NAVSIM~\cite{navsim} between DiffusionDrive~\cite{diffusiondrive} and our method, CaAD. While DiffusionDrive~\cite{diffusiondrive} often predicts ego trajectories that overlap with collision-related objects at future time steps, CaAD generates trajectories that avoid these objects while remaining close to the ground-truth driving behavior. The BEV visualization further shows that this difference comes from the predicted ego trajectory region at the same future time step, highlighting the ability of CaAD to produce safer and more interaction-aware plans in object-conflict scenarios.

\section{Discussions}
\myparagraph{Limitations.} Despite the strong overall closed-loop results, CaAD still struggles with rare social behaviors. In particular, we did not observe successful cases in which the ego vehicle reliably clears space for an approaching emergency vehicle, a failure mode that is also common in prior end-to-end driving frameworks~\cite{hipad,diver,hydranext}. More broadly, the gains are strongest in interaction-critical scenarios, while performance can still degrade when key agents are missed or when the scene contains rare conventions that are underrepresented in the training data. Our evaluation is also limited to camera-based simulation benchmarks.

\myparagraph{Social Impact.} 
CaAD can contribute to safer and more socially compliant autonomous driving by improving how end-to-end planners reason about interactions between the ego vehicle and surrounding agents. This is particularly beneficial in interaction-critical scenarios such as merging, yielding, overtaking, and intersection crossing, where reliable driving requires coordinated behavior among multiple traffic participants. More broadly, CaAD provides a positive step toward end-to-end autonomous driving systems that better capture real-world traffic dynamics and improves reliability for autonomous driving.

\end{document}